\documentclass[12pt]{article}
\usepackage{graphicx} 
\usepackage{amsmath}
\usepackage{lipsum}
\usepackage{pdfcomment}

\usepackage{tikz}
\usepackage{caption}
\usepackage{amsmath}
\usetikzlibrary{positioning}
\usetikzlibrary{arrows.meta, positioning}

\usepackage{xcolor} 
\usepackage{subcaption}
\usepackage[utf8]{inputenc}
\usepackage{longtable}
\usepackage{graphicx}
\usepackage{float}
\usepackage{booktabs}
\usepackage{float} 
\usepackage{geometry}
\usepackage{lineno}
\usepackage{authblk}
\usepackage{tcolorbox}
\usepackage{rotating}
\usepackage{booktabs}
\usepackage{threeparttable}
\usepackage{geometry} 
\usepackage{lscape}
\usepackage{siunitx}  
\usepackage{adjustbox} 
\usepackage{pgfplots}
\usepackage{caption}
\usepgfplotslibrary{groupplots}

\usepackage[numbers,sort&compress]{natbib}

\usepackage{hyperref}
\hypersetup{
  colorlinks=true,
  linkcolor=blue,       
  citecolor=blue,       
  urlcolor=blue         
}

\usepackage{cite}

\usepackage{algpseudocode}  
\usepackage[ruled,vlined]{algorithm2e}  

\usepackage{array}
\usepackage{multirow}

\usepackage{amssymb}
\usepackage{pifont}

\usepackage[colorinlistoftodos]{todonotes}

\usepackage{wrapfig}

\usepackage{cite}

\usepackage{amsthm}

\usepackage{titlesec}
\usepackage{titletoc}
\titleclass{\subsubsubsection}{straight}[\subsection]
\newcounter{subsubsubsection}[subsubsection]
\renewcommand\thesubsubsubsection{\thesubsubsection.\arabic{subsubsubsection}}
\titleformat{\subsubsubsection}{\normalfont\normalsize\bfseries}{\thesubsubsubsection}{1em}{}
\titlespacing*{\subsubsubsection}{0pt}{3.25ex plus 1ex minus .2ex}{1.5ex plus .2ex}
\titleformat*{\section}{\fontsize{11}{12}\selectfont\bfseries}
\titleformat*{\subsection}{\fontsize{11}{12}\selectfont\bfseries}
\titleformat*{\subsubsection}{\fontsize{11}{12}\selectfont\bfseries}
\titleformat*{\paragraph}{\fontsize{11}{12}\selectfont\bfseries}
\titleformat*{\subparagraph}{\fontsize{11}{12}\selectfont\bfseries}

\titlecontents{subsubsubsection}
[8em] 
{\small}
{\hyperlink{toc.\thecontentslabel}{\thecontentslabel.} }
{}
{\ \titlerule*[.5pc]{.}\contentspage}

\titlespacing*{\subsubsubsection}{0pt}{3.25ex plus 1ex minus .2ex}{1.5ex plus .2ex}

\usepackage{caption}
\usepackage{microtype}

\usepackage{enumitem}
\setlist{itemsep=0em}

\usepackage{fancyhdr} 
\usepackage{lastpage} 

\usepackage{listings}

\definecolor{codegreen}{rgb}{0,0.6,0}
\definecolor{codegray}{rgb}{0.5,0.5,0.5}
\definecolor{codepurple}{rgb}{0.58,0,0.82}
\definecolor{backcolour}{rgb}{0.95,0.95,0.92}

\lstdefinestyle{mystyle}{
    backgroundcolor=\color{backcolour},   
    commentstyle=\color{codegreen},
    keywordstyle=\color{magenta},
    numberstyle=\tiny\color{codegray},
    stringstyle=\color{codepurple},
    basicstyle=\ttfamily\footnotesize,
    breakatwhitespace=false,         
    breaklines=true,                 
    captionpos=b,                    
    keepspaces=true,                 
    numbers=left,                    
    numbersep=5pt,                  
    showspaces=false,                
    showstringspaces=false,
    showtabs=false,                  
    tabsize=2
}

\title{Text-to-Layout: A Generative Workflow for Drafting Architectural Floor Plans Using LLMs}

\author[1]{Jayakrishna Duggempudi}
\author[1]{Lu Gao}
\author[1]{Ahmed Senouci}
\author[2]{Zhe Han}
\author[3]{Yunpeng Zhang}

\affil[1]{Department of Civil and Environmental Engineering, University of Houston}
\affil[2]{Center for Transportation Research, The University of Texas at Austin}
\affil[3]{Department of Information Science Technology, University of Houston}

\date{}

\begin{document}

\maketitle

\abstract{This paper presents the development of an AI-powered workflow that uses Large Language Models (LLMs) to assist in drafting schematic architectural floor plans from natural language prompts. The proposed system interprets textual input to automatically generate layout options including walls, doors, windows, and furniture arrangements. It combines prompt engineering, a furniture placement refinement algorithm, and Python scripting to produce spatially coherent draft plans compatible with design tools such as Autodesk Revit. A case study of a mid-sized residential layout demonstrates the approach’s ability to generate functional and structured outputs with minimal manual effort. The workflow is designed for transparent replication, with all key prompt specifications documented to enable independent implementation by other researchers. In addition, the generated models preserve the full range of Revit-native parametric attributes required for direct integration into professional BIM processes. }

\vspace{0.5cm}
\noindent \textbf{Keywords}: Large Language Models, Architectural Design, Floor Plan Generation, Revit, Automation, Efficiency, Creativity, Building Information Modeling (BIM), Spatial Optimization, Artificial Intelligence in Design

\section{Introduction}
The development of automatic architectural design has gone through several phases. The earliest milestone is \citet{sutherland1964sketch}'s Sketchpad, which demonstrated that computers could manage and enforce geometric constraints through an interactive graphical interface. In the 1970s, rule-based systems emerged as a major paradigm. For example, \citet{stiny1971shape} introduced shape grammars. This formalism enables the recursive generation of architectural plans. It does so by applying production rules to geometric primitives. Later, a significant transition occurred during the 1980s and 1990s with the rise of commercial CAD software such as AutoCAD. While these tools were not automatic design systems themselves, they played a foundational role by formalizing geometry, topology, and object hierarchies in digital form \citep{eastman2018building}. Since the 2010s, data-driven generative methods have begun to transform architectural design. Machine learning techniques, particularly deep generative models, allow systems to learn from large datasets rather than relying solely on predefined rules. For example, \citet{chaillou2020archigan} employed generative adversarial networks (GANs) to convert site boundary inputs into detailed apartment layouts and interior configurations. 

Recently, various generative AI (GenAI) methods, such as variational autoencoders (VAEs), diffusion models, 3D generative models, and multimodal large language models, have been used for increasingly complex architectural and construction tasks \citep{li2025generative, sahraoui2025integrating,alwashah2025generative,madireddy2025large}. For example, \citet{hanafy2023artificial} investigates the impact of AI-based text to image tools such as DALL·E 2, Midjourney, and Stable Diffusion on creativity within architectural design processes. The text concludes that these tools improve ideation and visual exploration in conceptual design. However, they have technical limitations in precision tasks. \citet{onatayo2024generative} also noted that while generative AI shows promise in architectural design, its current uses are often limited to offering visual ideas or producing results that may lack functional detail, spatial logic, and compatibility with other design software. As a result, their outputs often require significant manual interpretation and rework before they can be meaningfully used in the design process.

This study explores how Generative AI, particularly LLMs, can support the automated drafting of early-stage architectural floor plans from natural language descriptions. It introduces a workflow that bridges language-to-layout translation, spatial feasibility checks, and compatibility with architectural modeling environments. Rather than generating finished models, the system helps architects start the design process by offering structured draft layouts, which supports early-stage ideation and functional spatial planning. The workflow is openly documented in this paper, and its outputs preserve complete Revit-native parametric attributes for direct use in BIM workflows without manual enrichment.


The remainder of this paper is organized as follows. Section 2 reviews related work on traditional and AI-driven methods for architectural design. Section 3 presents the methodology, including prompt structuring, data encoding, and the AI-to-Revit integration pipeline. Section 4 describes a case study that illustrates the proposed system in action and evaluates its performance. Finally, Section 5 concludes the paper and discusses potential directions for future research.

\section{Literature Review}

\subsection{Traditional Approaches to Floor Plan Generation}
Prior studies have approached floor plan generation through rule-based systems and evolutionary algorithms. Rule-based systems, such as shape grammars and L-systems, operate on predefined rules to generate designs that adhere to specific user-defined goals and constraints \citep{zeytin2024role}. These methods have been used to create designs within a particular style, such as Palladian villa ground plans, American bungalows, and traditional Marrakech houses, by defining elements like room sizes and semantic tags \citep{sonmez2018review}. However, their generation capability is bound by the underlying rules and representations, which can limit creative recombination \citep{ploennigs2024automating}. For example, \citet{ko2023architectural} reviews 589 studies on AI-driven architectural spatial layout planning and identifies key approaches and future research directions to support architects' creative roles. The authors observed that these methods often depend on empirical parameters. They also have limited adaptability to different formats. These methods frequently require substantial manual effort to adjust parameters and create detailed rules. 

As an alternative to rule-driven approaches, evolutionary algorithms, such as genetic algorithms and simulated annealing, have been introduced to treat design generation as an optimization problem.  These methods treat the design process as an optimization problem, iteratively searching for solutions that meet requirements \citep{gupta2023role}. For instance, \citet{munavalli2022dynamic} applied genetic algorithms to optimize hospital room layouts based on adjacency, circulation, and natural lighting. Despite these successes, evolutionary algorithms can struggle with multi-objective optimization problems where numerous complex and often competing goals need to be balanced \citep{puatruaușanu2024systematic}.

\subsection{Generative AI for Architectural Design}
GenAI refers to a class of AI models that can learn patterns from data and generate new content that mimics those patterns. Among the most prominent architectures in GenAI are Generative Adversarial Networks (GANs) and Diffusion Models, both of which have had a significant impact on spatial and layout-based design. GANs comprise two competing neural networks: a generator that creates new synthetic data, and a discriminator that distinguishes between real and generated data \citep{sukkar2024analytical}. This adversarial training process enables GANs to produce outputs that closely resemble real-world examples \citep{onatayo2024generative}. In architectural and spatial design context, GANs work through a competition between two neural networks: one generates data (like a floor plan), and the other evaluates how realistic it is. Over time, this adversarial process leads to highly realistic and creative outputs. GANs are employed to rapidly conceptualize and visualize design options \citep{amer2023architectural} by generating diverse design alternatives such as 2D floor plans \citep{zeytin2024role}, street view images \citep{zheng2021generative}, and even the geometry of curved surfaces \citep{lukovich2023artificial}. Furthermore, GANs have been applied to tasks like converting unstructured triangle meshes, classifying design objects, generating urban forms, and assisting in the structural design of high-rise buildings \citep{zeytin2024role}. Despite their power, training GANs can be computationally intensive, and image-based GANs may be inefficient for vectorized CAD data \citep{zheng2021generative}. They also face challenges like mode collapse, which limits output diversity, and may produce aesthetically pleasing but semantically nonsensical layouts if lacking true understanding \citep{hanafy2023artificial}.

Another significant architecture is the diffusion model, which underpins popular GenAI tools like Midjourney and DALL-E \citep{nichol2021glide}. These models operate by successively adding noise to an image until it becomes total noise, then learning the reverse process of denoising to generate coherent and intricate outputs \citep{ploennigs2024automating}. They excel in tasks such as denoising, inpainting, and super-resolution, and have revolutionized image generation \citep{ploennigs2024automating}. Diffusion models are widely used for text-to-image generation, allowing architects to rapidly produce high-quality architectural visualizations from textual prompts \citep{sukkar2024analytical}. Their applications extend to early conceptual design, where they can generate diverse design options, visualize abstract concepts, and refine design parameters \citep{tan2024using}. These tools offer a speed advantage by rapidly visualizing ideas, freeing architects to focus on strategic decisions and client communication \citep{lee2024generative}. However, diffusion models are primarily 2D image-generating tools and may not fully account for 3D spatial functionality. These models rely on carefully designed text prompts. Designers must clearly articulate their concepts. However, these models may generate visually pleasing but spatially unrealistic images. This occurs due to their limited understanding of architectural function and spatial organization \citep{shi2024generative}.

\subsection{Prompt to BIM Model}

Current tools capable of generating architectural design schemes from natural language or quasi-natural language prompts and integrating them with Revit include Arkio AI, TestFit, Hypar, Autodesk Forma, Planner 5D AI, BIMLOGIQ Copilot, BIMIL AI Helper, BIM-GPT, NADIA, and LLM4DESIGN. Arkio AI transforms prompts into conceptual massing models. These models are synchronized to Revit and users can derive walls, floors, and roofs from these models. However, the initial output does not include full parametric details such as element types, wall thicknesses, materials, story elevations, and constraint definitions \citep{ArkioRevitPlugin2024}. TestFit translates prompts, typically structured parameters rather than free, form text, into typology-specific building and parking layouts. Through its Revit plugin, these layouts can be pushed into Revit containing partial parametric information (e.g., room zoning, number of floors, and certain dimensional constraints), but still require the addition of detailed family definitions, element properties, and material data in Revit \citep{TestFitRevitIntegration2025}. Hypar converts prompts into space- and element-level BIM models that can be imported into Revit as natively editable elements, which often preserves categories, dimensions, story levels, and some materials. However, free-form text typically needs to be reformulated into predefined templates, and the platform is commercial and closed-source \citep{HyparRevitPlugin2025}. Autodesk Forma uses parameter-based inputs or limited natural language prompts to produce early-stage site layouts and performance-optimized building forms, which are exported as IFC or DWG files for import into Revit. The imported geometry lacks native BIM families, types, materials, and constraints, and Forma is commercial and closed-source \citep{AutodeskFormaIntegration2025}. Planner 5D AI converts prompts into 2D floor plans and 3D interior layouts. However, these exports are purely geometric and contain no BIM metadata such as element types, dimensions, materials, or elevation data, and the tool is not open-source \citep{Planner5DExportFormats2024}. BIMLOGIQ Copilot operates as an AI-assisted automation plugin within Revit that executes natural language commands to perform modeling and documentation tasks, such as creating views, modifying element parameters, and generating schedules. While it can create or update elements, it does not generate complete architectural models from scratch \citep{BIMLOGIQCopilot2025}. BIMIL AI Helper functions as an AI-assisted tool within Revit that uses natural language to perform modeling and validation tasks. It is commercial and closed-source. It can update or add elements while preserving Revit's native parametric information. However, it does not create complete architectural models from scratch \citep{BIMILAIHelper2024}. BIM-GPT is a prompt-based virtual assistant framework for BIM information retrieval that integrates BIM data with GPT models. Its primary function is focused on information retrieval rather than generating architectural models \citep{BIMGPT2023}. NADIA is a research prototype that assists in generating architectural detailing, such as exterior wall assemblies, from natural language descriptions. It interprets textual input to derive design intent and produces detailed construction layer information and node specifications that can be integrated with BIM environments \citep{NADIA2024}. LLM4DESIGN is a multimodal system that uses large language models, retrieval-augmented generation, and visual language models to generate conceptual architectural and environmental designs from natural language prompts. While it can produce visual layouts and annotated diagrams, it does not generate Revit-native models with full parametric attributes \citep{LLM4DESIGN2024}. Across all tools, none are open-source and only Arkio AI, TestFit, and Hypar provide a relatively direct prompt-to-Revit workflow. Even in these cases, the resulting models generally remain at a conceptual or partially parameterized level, which requires further enrichment within Revit to reach construction-ready BIM standards.

\subsection{Limitations and Research Gaps}



While AI technologies in architectural design have advanced significantly, most existing tools capable of generating design outputs are commercial and closed-source, with no mature open-source alternatives available. Furthermore, even when these tools integrate with professional CAD and BIM environments such as Revit, the generated results often lack the full range of Revit-native parametric information. For example, outputs may consist of conceptual massing or partial geometric models without complete element types, material specifications, and constraint definitions, which requires substantial manual enrichment before they become fully functional BIM models.

To address these limitations, this study makes two main contributions: (1) it creates an AI process that converts natural language descriptions into functional, semantically coherent, and BIM-compatible 3D floor plans. The paper includes all necessary prompts to ensure reproducibility and adaptability for other researchers; and (2) it produces Revit-native models enriched with complete parametric information, including element types, material specifications, wall thicknesses, story elevations, and constraint definitions. This enables immediate use in professional BIM environments without the need for manual enrichment.

\section{Methodology}

Architectural design often involves labor-intensive and repetitive tasks such as drafting floor plans, which can be time-consuming and prone to inefficiencies. This study proposes an AI-assisted workflow that automatically generates draft floor plan geometry from natural language prompts (e.g., “a three-bedroom apartment with a spacious living room”) and translates this data into 3D building information models using Autodesk Revit.

\subsection{Workflow Overview}
\begin{figure}[H]
    \centering
    \includegraphics[width=0.6\linewidth]{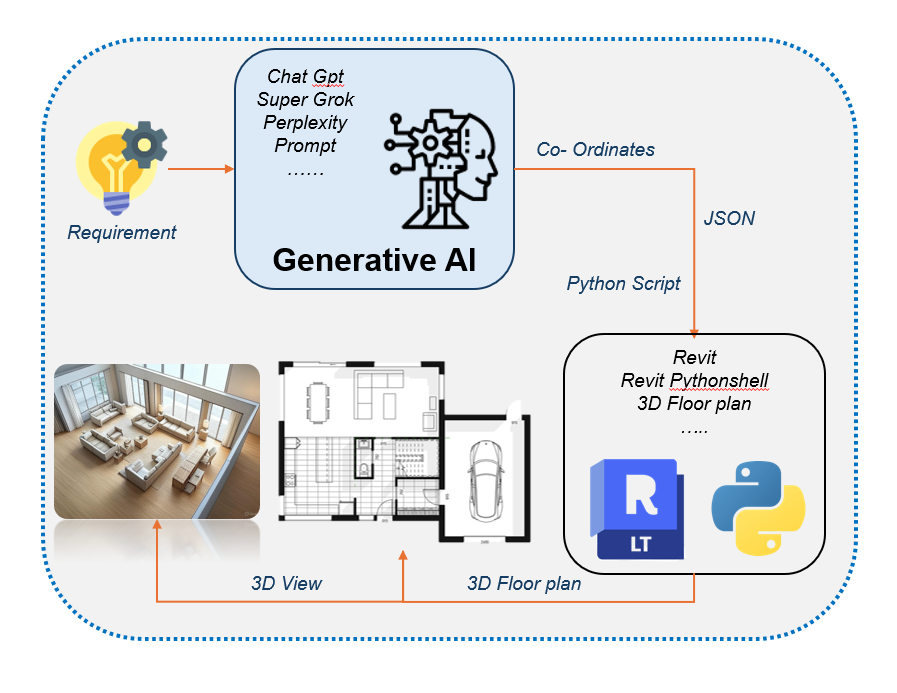}
    \caption{Workflow from Prompt to BIM Model}
    \label{fig:workflow}
\end{figure}

The workflow of the proposed system is illustrated in Figure \ref{fig:workflow}. This diagram demonstrates how GenAI models process textual inputs to produce spatial coordinates. These coordinates are then converted into 3D models compatible with BIM software, using a Revit-Python integration layer. The pipeline begins with user-defined requirements in natural language, which are processed by a large language model to produce a structured JSON file containing coordinates for architectural elements such as walls, doors, windows, and furniture. A custom Python script then interprets this file and programmatically constructs a corresponding 3D model in Autodesk Revit. This automated process significantly reduces manual drafting efforts while maintaining spatial logic and functional clarity.

We evaluated several LLMs, including GPT-4o, GPT-o4-mini, GPT-o4-mini-high, Claude 2.1, and Gemini 2.5 Pro. Among these, GPT-4o consistently produced the most reliable and well-structured outputs for our task-specific prompts. We use GPT-4o with default parameters, and provide example prompts, generated layout JSON, and Revit integration scripts in the supplementary material to support reproducibility.

In our experiments, the entire workflow (from prompt submission to Revit model generation) typically completes within 10 minutes on a laptop (Intel Core i7 CPU, 32GB RAM). The language model response and JSON parsing take approximately 2-3 minutes seconds, while Revit automation via Python scripts takes another 7-8 minutes depending on model complexity.

To further clarify the implementation, the following subsections detail how user-defined prompts are structured, how the AI system generates spatial data, and how this data is interpreted and refined through Revit scripting. This elaboration bridges the conceptual workflow with the technical execution of the system.

\subsection{Prompt structuring and data encoding}

The workflow begins with a single, well-structured natural-language prompt. Figure \ref{fig:prompt} shows the (abbreviated) prompt issued to the model:

\begin{figure}[H]
\centering
\begin{lstlisting}[basicstyle=\ttfamily\footnotesize, frame=single]
Generate JSON for a 30x40 ft single-storey floor plan.
Walls: exterior + interior for LivingHall, Kitchen, OfficeRoom,
       Bedroom (with attached toilet)
Doors: on walls, ensure connectivity, no overlaps
Windows: on exterior walls, no door overlaps
Furniture by room
  (LivingHall: Sofa, TVUnit;
   OfficeRoom: Sofa, OfficeDesk;
   Bedroom: Bed, Wardrobe;
   Kitchen: DiningTable, Bench)
use centre [x,y,0], no overlaps
\end{lstlisting}
\caption{Example natural-language prompt used to generate layout}
\label{fig:prompt}
\end{figure}

Because the natural-language prompt above is intentionally concise, we expand  each implicit instruction into an explicit, machine-readable specification (Table \ref{tab:prompt-structure}).  The table therefore represents the full prompt that tells the AI exactly how every architectural component should be represented in its JSON output.

\begin{table}[H]
\centering
\footnotesize
\caption{Expanded prompt specification: every element the model must output}
\label{tab:prompt-structure}
\renewcommand{\arraystretch}{1.15}
\begin{tabular}{|p{3cm}|p{9.2cm}|}
\hline
\textbf{Directive} & \textbf{Expanded instruction given to the model} \\
\hline
\texttt{walls} &
Return an array \texttt{walls[]} where each item has keys
\texttt{start} [x,y,0] and \texttt{end} [x,y,0].  
Exterior walls form a 30×40 ft rectangle; interior walls subdivide
the space into the four rooms named in the prompt.            \\ \hline
\texttt{doors} &
Return an array \texttt{doors[]} with the \texttt{start}/\texttt{end}
format above.  Place at least one exterior door to the LivingHall and
interior doors that connect all rooms; no door may overlap a window.  \\ \hline
\texttt{windows} &
Return an array \texttt{windows[]} on exterior walls only, in the same
\texttt{start}/\texttt{end} format; each window must be at least 2 ft from
any door edge.                                                   \\ \hline
\texttt{furniture} &
Return an object \texttt{furniture\{\}} whose keys are room names.
Each value is an array of items with  
\{\texttt{name}, \texttt{position} [x,y,0]\}.  
All furniture must respect a clearance $\ge$ 1 ft from walls, doors, windows,
and other furniture.                                           \\ \hline
\end{tabular}
\end{table}


Figure~\ref{fig:full_prompt} shows the exact prompt used to generate the final layout. Eight different prompts were tested in total. Each prompt varied in structure, detail, and formatting instructions. 

\begin{figure}[H]
\centering
\begin{lstlisting}[
    basicstyle=\ttfamily\small,
    frame=single,
    label={lst:complete-prompt},
    breaklines=true,
    breakatwhitespace=true,
    columns=fullflexible,
    postbreak=\mbox{\textcolor{gray}{$\hookrightarrow$}\space}
]
Generate a JSON object representing a floor plan for a single-story building 
with overall dimensions of 30 feet in width (x-axis: 0 to 30) and 40 feet in 
length (y-axis: 0 to 40). The output must be a JSON object with four top-level 
keys: "walls", "doors", "windows", and "Furniture".

walls: Provide an array of objects where each object represents a wall segment 
with "start" and "end" coordinates in the format [x, y, 0]. The exterior walls 
must form a 30x40 ft rectangle. Include interior walls to define the following 
rooms: LivingHall, Kitchen, OfficeRoom, and Bedroom with an attached toilet. 
The AI should decide the placement and dimensions of these rooms.

doors: List each door with "start" and "end" coordinates. Place doors on wall 
segments with no overlap. Ensure logical connectivity between rooms and include 
at least one exterior entry door.

windows: Provide an array of "start" and "end" coordinates. Place only on 
exterior walls, avoiding any overlap with doors.

Furniture: For each room, include furniture as objects with "name" and 
"position" fields. Position is [x, y, 0], representing the center. 
- LivingHall: Sofa, TVUnit
- OfficeRoom: Sofa, OfficeDesk
- Bedroom: Bed, Wardrobe
- Kitchen: DiningTable, Bench

Assume standard furniture dimensions and ensure no overlaps with walls, 
doors, or other furniture. All components must lie within the defined 
room boundaries.

Note: Avoid unnecessary text or metadata. Output should be a clean JSON 
object only.
\end{lstlisting}
\caption{Full Prompt for Generating Floor Plan JSON Layout}
\label{fig:full_prompt}
\end{figure}

Guided by the expanded prompt specification, the generative model produces a JSON file that encodes the entire layout. Linear elements (walls, doors, windows) are stored as start/end coordinate pairs, while furniture items are stored as centre-point coordinates grouped by room. This structured output, which conforms exactly to the schema in Table~\ref{tab:prompt-structure}, serves as a digital blueprint for the subsequent Revit automation stage. An excerpt of the generated JSON for \texttt{walls} is shown in Figure \ref{fig:json-walls}.

\begin{figure}[H]
\centering
\begin{lstlisting}[basicstyle=\ttfamily\small, frame=single]
"walls": [
    {"start": [0,0,0], "end": [30,0,0]},
    {"start": [30,0,0], "end": [30,40,0]},
    {"start": [30,40,0], "end": [0,40,0]},
    {"start": [0,40,0], "end": [0,0,0]},
    {"start": [0,20,0], "end": [30,20,0]},
    {"start": [15,20,0], "end": [15,40,0]},
    {"start": [15,30,0], "end": [30,30,0]},
    {"start": [5,20,0], "end": [5,40,0]}
]
\end{lstlisting}
\caption{JSON structure for wall definitions in a 30x40 ft layout}
\label{fig:json-walls}
\end{figure}

To convert the JSON-based layout into an executable Revit model, we use a second-stage prompt that instructs the LLM to generate Python code compatible with the Revit API. This prompt, shown in Figure~\ref{fig:wall-prompt}, focuses specifically on creating wall elements from coordinate data defined in the \texttt{"walls"} section of the JSON file. It specifies required API functions, transaction structure, and output constraints to ensure script validity and direct usability within the Revit Python Shell.

\begin{figure}[H]
\centering
\begin{minipage}{0.95\textwidth}
\begin{lstlisting}[basicstyle=\ttfamily\small, frame=single]
You are a Revit Python expert. Given the variable `data` that stores a JSON object with a top-level key "walls", where each element has "start" and "end" coordinates in the format [x, y, 0], generate a Python script that creates all the walls in Autodesk Revit.

Requirements:
- Import the necessary Revit API classes (e.g., Autodesk.Revit.DB).
- Begin a Transaction.
- For each wall segment:
  - Create a Line using Line.CreateBound(XYZ(x1, y1, 0), XYZ(x2, y2, 0)).
  - Use Wall.Create(doc, line, wallType.Id, level.Id, 10, 0, False, False) to place a 10-ft-high basic wall.
- Use the first available wall type (wallType) and the first level (level) in the document.
- Commit the Transaction.
- Do NOT generate doors, windows, or furniture.
- Output only executable Python code. Do not include any explanations, comments, or markdown.
\end{lstlisting}
\end{minipage}
\caption{Prompt used to generate a Revit Python script from JSON wall data}
\label{fig:wall-prompt}
\end{figure}

\subsection{Greedy Algorithm for Furniture Placement Adjustment}
\label{sec:algorithm}

The LLMs approach discussed earlier can create initial layouts with basic room structures and furniture positions. However, their outputs may not always place items in expected or desired locations. For example, they might not align a bed against a wall or center a table in a room. In Revit, when using Python through the API, furniture cannot be drawn directly like walls but must be inserted as predefined Family instances, with only a single center point specified for placement. Because different Families have varying footprints, naive placement can lead to overlaps or collisions with surrounding elements. These spatial inaccuracies can affect both the functionality and realism of the layout. To address this, we introduce a greedy algorithm that incrementally adjusts the AI-generated positions, evaluating clearances and moving furniture toward intended placements such as wall-adjacent or centrally located positions until all spatial constraints are satisfied. Because the adjustment process is computationally inexpensive, we did not explore more complex optimization strategies. Instead, this efficient and interpretable method serves as a practical solution to refine layout fidelity and ensure clash-free arrangements with minimal overhead.



Let $\mathcal{F}=\{f_1,f_2,\dots,f_m\}$ be the set of furniture items.
For a given item $f_i$ we write $p_i^{(0)}\in\mathbb{R}^2$ for the
AI-generated centre point, $\text{shape}(f_i)$ for its rectangular
footprint, $\Omega_i$ for the polygonal boundary of the room that
contains $f_i$, and $\mathcal{W}_i$ for the set of wall segments that
constitute $\partial\Omega_i$.

A candidate position $p$ induces an axis-aligned bounding box $B_i(p)$ obtained by translating $\text{shape}(f_i)$ so that its centre coincides with $p$.  For items such as beds or sofas that are normally placed against a wall,
we define the headboard edge to be the footprint edge with the largest $y$-extent, although any consistent convention is acceptable. 

If $A$ and $B$ are two rectangles, the function $\text{dist}(A,B)$ returns the Euclidean distance between their closest
points and equals zero when they overlap. Throughout our residential tests we set the minimum walking clearance to $\delta=1$\,ft, but this value can be increased (for example, to 1.5 ft for accessible routes).

Placing $f_i$ at $p$ is called \emph{feasible} when the following three criteria are satisfied simultaneously:

\begin{align}
\text{Feasible}(f_i,p)=1 \;\Longleftrightarrow\;
\begin{cases}
\text{(a)} & B_i(p)\subset\text{Interior}(\Omega_i),\\[3pt]
\text{(b)} & \displaystyle
            \min_{o\in\mathcal{O}}\text{dist}\bigl(B_i(p),B_o\bigr)\ge\delta,\\[9pt]
\text{(c)} & \exists\,w\in\mathcal{W}_i\text{ such that the headboard edge of }B_i(p)\parallel w.
\end{cases}
\end{align}

Condition (a) ensures that the footprint remains within the room's boundaries. Condition (b) guarantees a collision-free buffer of at least $\delta$ between the new footprint and every obstacle $B_o\,(o\in\mathcal{O})$, where $\mathcal{O}$ contains all walls, doors, windows and previously placed furniture.  
Condition (c) enforces the usual interior-design preference that certain
items rest flush against a wall; for freestanding objects such as dining
tables the test is disabled.

These three constraints collectively ensure that each furniture item remains legally within its designated room, allows for sufficient circulation space, and, where applicable, aligns neatly with a wall. Items that violate any criterion are incrementally nudged towards the nearest wall by the greedy algorithm described in the following sub-section until feasibility is achieved. 

If a furniture item fails to meet any of these constraints at its initial AI-generated position,
a greedy wall-seeking algorithm is applied to incrementally adjust its placement until all conditions are satisfied. In this algorithm, we iteratively adjust each piece towards the nearest viable wall until the feasibility test is satisfied.

\begin{algorithm}[H]
\small
\caption{GreedyWallPlacement($f_i$)}
\KwIn{
  Initial position $p \leftarrow p_i^{(0)}$;\\
  Furniture footprint $\text{Shape}(f_i)$;\\
  Wall set $\mathcal{W}_i$ for room $\Omega_i$;\\
  Current occupied space $\mathcal{O}$
}
\KwOut{Final feasible position $p_i^\ast$}

\While{\textnormal{Feasible}$(f_i, p) = 0$}{
  Compute direction ${\boldsymbol v}_{\text{wall}}(p)$ toward nearest wall in $\mathcal{W}_i$;\\
  Move $p \leftarrow p + \lambda \cdot {\boldsymbol v}_{\text{wall}}(p)$;
}
$p_i^\ast \leftarrow p$;\\
Update global occupancy: $\mathcal{O} \leftarrow \mathcal{O} \cup B_i(p_i^\ast)$;
\end{algorithm}

Here ${\boldsymbol v}_{\text{wall}}(p)$ denotes the unit vector pointing from the current furniture position $p$ toward the nearest wall segment in $\mathcal{W}_i$, and $\lambda$ is a small step size (e.g., $0.5\,$ft). This incremental movement mimics a gradient-descent-like behavior, guiding the item toward a feasible position without exhaustively searching the space.

The algorithm operates sequentially over all furniture items $f_i \in \mathcal{F}$. For each item, it starts at the AI-suggested initial location $p_i^{(0)}$ and repeatedly nudges the item toward the nearest wall until all placement constraints are satisfied. At each step, it checks whether the new location meets three critical conditions: (1) it remains inside the room boundary, (2) it avoids collisions with existing objects (tracked in the global occupancy set $\mathcal{O}$), and (3) it aligns with a wall when applicable (e.g., for beds or sofas). Once a valid placement is found, the item's bounding box $B_i(p)$ is added to $\mathcal{O}$ to prevent future overlaps. This method is straightforward and efficient. Unlike brute-force approaches that assess a dense set of candidates, the greedy wall-seeking algorithm utilizes spatial intuition, which recognizes that furniture is usually placed against walls. It quickly converges to a practical configuration with minimal computation.

\section{Case Study}

This case study illustrates the development and evaluation of an AI-assisted system for generating residential floor plans. The goal was to design a 30×40 ft single-story building that includes a Living Hall, Kitchen, Office Room, and Bedroom with an attached toilet. Using generative AI to translate textual prompts into spatial data, we automated the creation of a Revit model via a Python script. The purpose of this experiment was to test the effectiveness of our workflow in producing functionally and spatially coherent layouts without manual drafting.

\begin{figure}[H]
    \centering
    \includegraphics[width=0.6\linewidth]{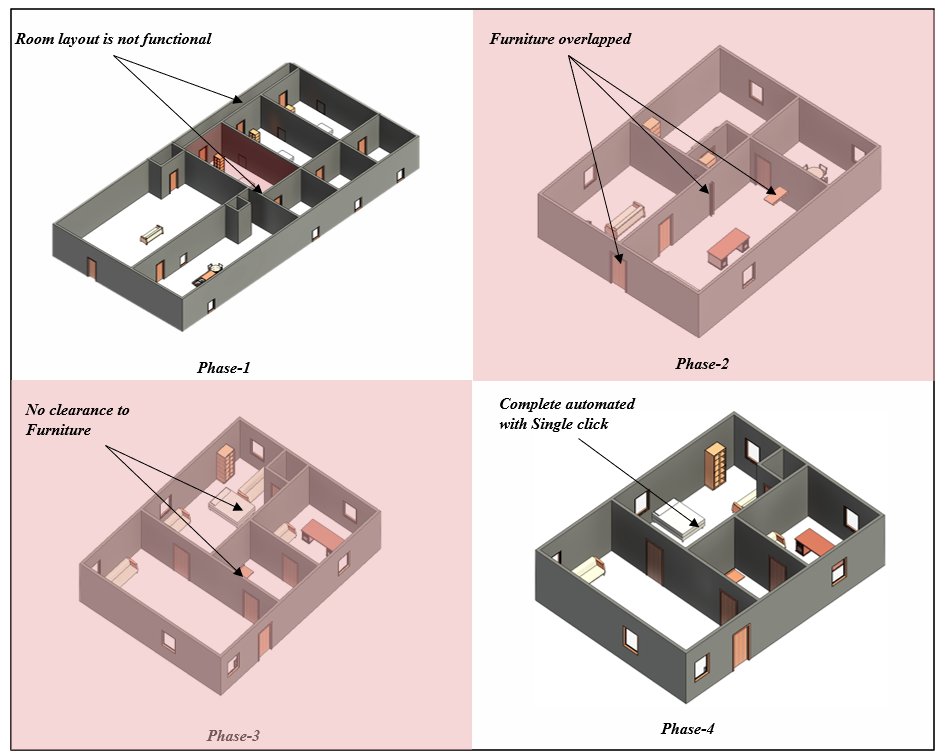}
    \caption{Examples of layout issues observed during early experimentation.}
    \label{fig:house}
\end{figure}

Figure~\ref{fig:house} highlights several issues identified during the early experimental phase. While the AI-generated JSON layouts were able to produce basic spatial configurations, the resulting plans exhibited a number of architectural and functional flaws. These included: rooms with ambiguous boundaries or improper dimensions, furniture overlapping with walls or other elements, insufficient circulation space between items, and illogical door placements (e.g., doors swinging into obstacles or placed in corners).

To address these problems, we developed a set of post-processing enhancements within the Revit Python script, focusing on spatial logic correction and layout refinement. The improvements included:

\begin{itemize}
    
    \item Each room is evaluated based on expected furniture content (e.g., bedrooms must include a bed and wardrobe). If required items are missing or the area is insufficient, the system flags the space for redesign or merging with adjacent zones.
    
    \item The system checks that doors and windows are positioned along valid wall segments and do not interfere with adjacent elements. It also ensures that door swing paths are unobstructed and aligned with practical access routes.
    
    \item A grid-based pathfinding algorithm is used to assess connectivity between key functional spaces. If any pathway is blocked or too narrow, the layout is modified to ensure clear and logical circulation.
\end{itemize}

These rule-based corrections and geometric analyses significantly improved the quality and usability of the generated floor plans. By combining generative AI with a robust Revit-based adjustment mechanism, we were able to produce architectural models that better align with both functional requirements and design intent.

Figure \ref{fig:refinement} presents a comparative visual overview of six AI-generated 3D architectural layouts, produced using variations of natural language prompts during early-stage experimentation. Each model rendered in a transparent wireframe with magenta highlights captures different spatial compositions and furniture arrangements, which reflects the evolving effectiveness of prompt engineering. Early outputs, such as those in the top row, often exhibited issues like excessive partitioning, awkward circulation, or cluttered furnishing. For example, several layouts lacked logical room adjacencies or included furniture blocking doors. These errors stemmed from under-specified prompts or missing constraints. As the prompt syntax became more structured, it included explicit rules for room connectivity, furniture clearance, and wall alignment. This led to clearer zoning, more efficient layouts, and fewer spatial conflicts in the models. This figure illustrates not only the generative model’s sensitivity to prompt detail but also the iterative refinement process required to arrive at usable results. The progression from top-left to bottom-right demonstrates the tangible improvements achieved through hundreds of testing cycles. By systematically refining prompt content, such as specifying room functions, component positions, and spacing logic, we improved the AI's ability to generate functional, aesthetically cohesive, and BIM-compatible layouts.

\begin{figure}[H]
    \centering
    \includegraphics[width=0.6\linewidth]{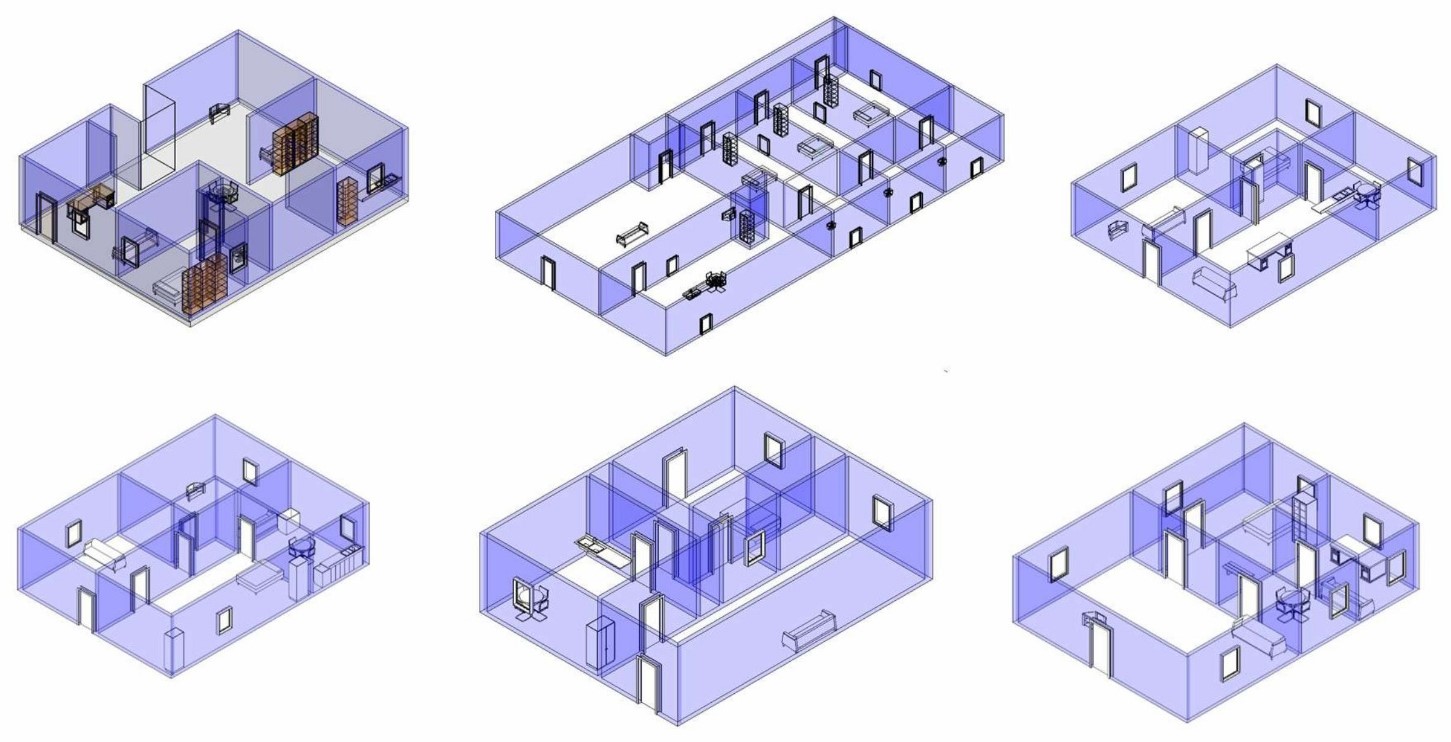}
    \caption{Evolution of AI-generated 3D floor plans through prompt refinement. (1–6 ordered left to right, top to bottom.) (1) Over-partitioned layout with poor circulation. (2) Unzoned space lacking clear functions. (3) Overcrowded and illogical adjacency. (4) Improved zoning but still misaligned furniture. (5) Better layout with basic clearance rules. (6) Functional and coherent design with optimized room placement.}
    \label{fig:refinement}
    \label{fig:refinement}
\end{figure}

\begin{figure}[H]
    \centering
    \includegraphics[width=0.8\linewidth]{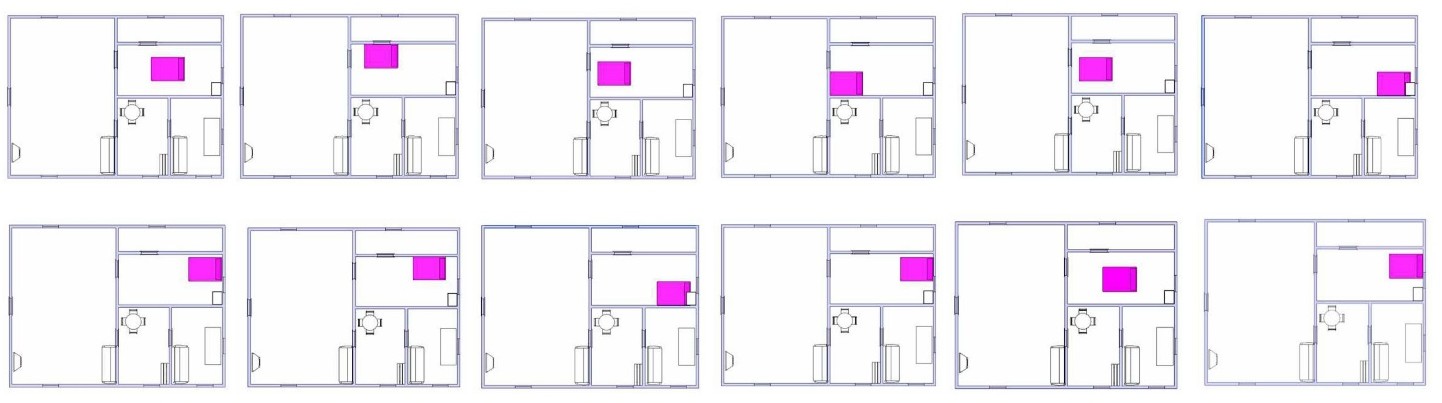}
    \caption{Furniture Iterations using the Python code}
    \label{fig:iteration}
\end{figure}

Figure \ref{fig:iteration} illustrates the use of the greedy algorithm to automatically optimize the placement of a bed within a room, showcasing both the iterative refinement process and the final result. The figure includes multiple top-down views and a final 3D model, visually documenting how the algorithm systematically adjusts the bed’s position to meet spatial requirements. The process begins with an initial bed location derived from the AI-generated JSON layout. However, this default position may suffer from issues such as overlap with walls or furniture, insufficient clearance, or misalignment. To resolve this, the greedy algorithm incrementally nudges the bed toward the nearest feasible wall. After each small movement, it performs a feasibility check to ensure that the bed remains entirely within the room, does not overlap with other objects, and has its headboard aligned with a wall. The figure shows several intermediate placements evaluated during the process. The final 3D model demonstrates a successful outcome: the bed is positioned against the top wall, with the headboard properly aligned, and the other three sides maintaining sufficient clearance from all surrounding elements.

\begin{figure}[H]
    \centering
    \includegraphics[width=0.8\linewidth]{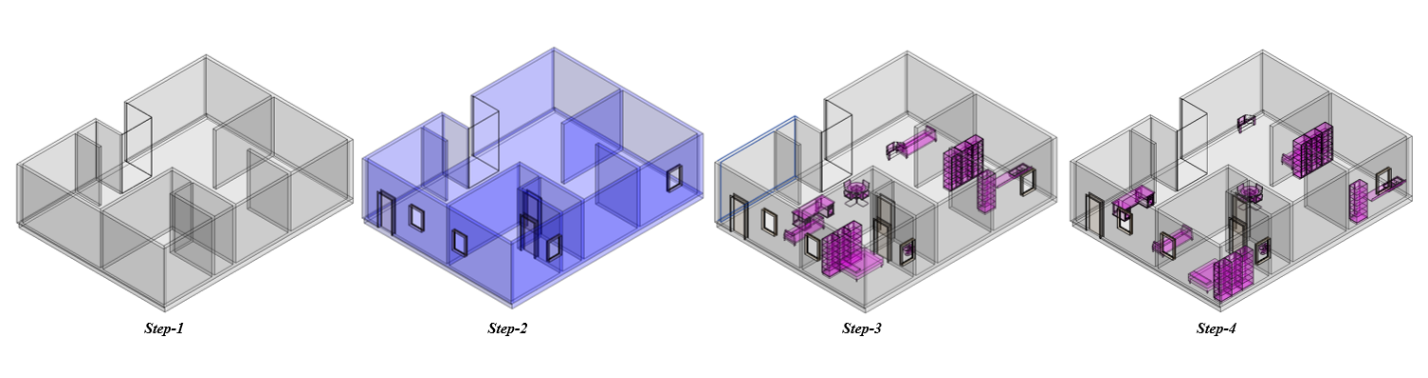}
    \caption{Automation steps for AI-assisted layout generation using Revit Python scripts. (Steps 1–4, ordered left to right): (1) Wall structures are generated from JSON inputs. (2) Doors and windows are automatically inserted. (3) Furniture items are placed based on initial coordinates. (4) Furniture positions are refined for spacing, clearance, and usability.}
    \label{fig:all}
\end{figure}

Figure \ref{fig:all} illustrates the complete automation pipeline used in this case study, showcasing how a Revit model is progressively generated from JSON-based input through four sequential steps. 

The process shows how Python scripting, run in the Revit Python Shell, can help create and improve an interior layout from basic structure to optimized furniture placement. In the first panel (Step-1), the process begins with the generation of the wall structure. Gray wireframe lines define a rectangular boundary, partitioned internally to form multiple rooms and open areas. At this point, there are no doors, windows, or furniture. The design features only the basic architectural structure. A custom Python script parses wall specifications such as position and dimension from the JSON input to generate this layout. The isometric view highlights the foundational spatial configuration established in this step. In the second panel (Step-2), doors and windows are added to the existing structure. Blue shaded regions appear along the gray wireframe walls, marking the locations of doorways and windows on both exterior and interior walls. These components are automatically placed using another Python script, which reads from the same JSON file to determine their types, sizes, and coordinates. The consistent isometric perspective shows how these openings are integrated seamlessly into the structure. In the third panel (Step-3), furniture objects are introduced. Purple shaded shapes indicate various items. These include beds, sofas, desks, and tables. These items are located in different rooms. The previously placed blue doors and windows remain visible, while the purple furniture varies in size and form to reflect the diversity of elements in the interior layout. The script uses position data from the JSON file to place each item accordingly. The fourth panel (Step-4) represents the refinement phase. Compared to Step-3, the furniture items are now more organized, with consistent clearance maintained from walls, doors, and other furniture. This improvement is the result of an automated adjustment algorithm implemented in Python, which tests and modifies each furniture position for functional alignment and ergonomic spacing. While the doors and windows remain fixed, the updated furniture layout appears more deliberate and optimized for usability.

\section{Conclusions}
This paper presents an automated workflow for drafting architectural floor plans from natural language prompts. The proposed system converts textual descriptions into structured layout data using LLMs. This data is then processed into draft-level geometry through JSON encoding and Python scripting within a Revit environment. A greedy wall-seeking algorithm was proposed to adjust furniture placements based on spatial constraints. Through prototype experiments and a residential case study, the workflow demonstrates its ability to produce coherent, furniture-aware layouts with minimal manual input. 

Unlike most existing Prompt-to-BIM tools that are commercial and closed-source, this approach offers an openly documented workflow and produces outputs in a flexible Revit-native format that can be customized to include all required parametric attributes for immediate use in BIM environments. Building on this capability, the contribution of this paper lies in providing an openly documented workflow that ensures transparency and enables reproducibility without reliance on proprietary systems. It also supports community-driven adaptation, which lowers integration barriers and extends the method's accessibility beyond any single commercial product.

Based on experiments and case studies, the following key insights were identified.

\begin{enumerate}
    \item Structured natural language prompts enhance the clarity and interpretability of AI-generated layouts, which provides a solid basis for subsequent modeling.
    
    \item Use of JSON as an intermediate representation decouples design intent from geometric realization, which enables smooth integration between generative models and BIM software and enhancing cross-platform compatibility.
    
    \item The greedy refinement algorithm for furniture placement effectively resolves spatial conflicts and improves both the functionality and aesthetics of the final design. 

\end{enumerate}

Despite these contributions, several limitations remain: (1) The test environment was constrained to simple, single-story rectangular layouts. The system's applicability to more complex geometries or multi-level buildings remains to be explored. (2) The current approach depends on manually crafted prompt templates, which may limit scalability and generalizability; future work could investigate automated or adaptive prompt generation methods. (3) The refinement mechanism relies on a greedy search algorithm. While effective for small-scale furniture layouts, this method may encounter performance challenges when applied to larger or more densely populated spatial configurations. (4) This proposed approach generates only an initial draft plan. It does not consider performance metrics such as space efficiency, circulation quality, and overlap rate. These factors should be addressed in future research. (5) The authors of this paper include trained and licensed architects. However, we did not have an external expert independently evaluate the proposed work. (6). In this study, we only proposed the workflow within Revit. To adapt this workflow for other platforms, the only required modification is the prompt, as illustrated in Figure \ref{fig:wall-prompt}, where platform-specific instructions should be provided to guide script generation. (7). This study did not incorporate building code compliance, accessibility standards, or other regulatory constraints into the case study. However, such requirements can be integrated into the JSON generation prompt (e.g., the one shown in Figure~\ref{fig:full_prompt}) by specifying them explicitly. Incorporating these constraints represents a promising direction for future research.

\bibliographystyle{unsrtnat}
\bibliography{references}

\end{document}